\documentclass{article}

\PassOptionsToPackage{numbers, compress}{natbib}
\usepackage[preprint]{neurips_2022}

\usepackage[utf8]{inputenc} % allow utf-8 input
\usepackage[T1]{fontenc}    % use 8-bit T1 fonts
\usepackage{hyperref}       % hyperlinks
\usepackage{url}            % simple URL typesetting
\usepackage{booktabs}       % professional-quality tables
\usepackage{amsfonts}       % blackboard math symbols
\usepackage{nicefrac}       % compact symbols for 1/2, etc.
\usepackage{microtype}      % microtypography
\usepackage{xcolor}         % colors

\usepackage{algorithm}
\usepackage{algorithmic}

\usepackage{nicefrac}
\usepackage{microtype}
\usepackage{graphicx}
\usepackage{caption}
\usepackage{subcaption}
\usepackage{tikz-cd}
\usetikzlibrary{arrows}
\usepackage{comment}
\usepackage{amssymb}
\usepackage{tikz}
\usepackage{tikz-qtree,tikz-qtree-compat}
\usetikzlibrary{positioning}

\usepackage{amsmath}
\usepackage{amsthm}
\usepackage{tikz}
\usepackage{tikz-qtree,tikz-qtree-compat}

\makeatletter
\newtheorem*{rep@theorem}{\rep@title}
\newcommand{\newreptheorem}[2]{%
\newenvironment{rep#1}[1]{%
 \def\rep@title{#2 \ref{##1}}%
 \begin{rep@theorem}}%
 {\end{rep@theorem}}}
\makeatother

\newtheorem{theorem}{Theorem}
\newreptheorem{theorem}{Theorem}

\usetikzlibrary{positioning}
\newtheorem{definition}[theorem]{Definition}

\usepackage{stackengine}

\title{Unit Selection: Case Study and Comparison with A/B Test Heuristic}

\author{%
  Ang Li, Judea Pearl\\
  Cognitive Systems Laboratory, Department of Computer Science\\
  University of California Los Angeles\\
  Los Angeles, CA 90095 \\
  \texttt{\{angli,judea\}@cs.ucla.edu} \\
}

\begin{document}

\maketitle

\begin{abstract}
The unit selection problem defined by Li and Pearl identifies individuals who have desired counterfactual behavior patterns, for example, individuals who would respond positively if encouraged and would not otherwise. Li and Pearl showed by example that their unit selection model is beyond the A/B test heuristics. In this paper, we reveal the essence of the A/B test heuristics, which are exceptional cases of the benefit function defined by Li and Pearl. Furthermore, We provided more simulated use cases of Li-Pearl's unit selection model to help decision-makers apply their model correctly, explaining that A/B test heuristics are generally problematic.
\end{abstract}

\section{Introduction}
The unit selection problem defined by Li and Pearl is encountered in several fields, such as business, social science, health science, and economics. For example, decision-makers in customer relationship management want to identify customers who would buy purchases if there is an enticement and would not otherwise \cite{berson1999building, hung2006applying,lejeune2001measuring, tsai2009customer}. For another example, large companies are interested in customers who would visit their website if there the website is prompted by an online advertisement and would not otherwise \cite{bottou2013counterfactual, li2020training,li2014counterfactual, sun2015causal, yan2009much}. The challenge of identifying such customers stems from the fact that the desired behaviors are defined counterfactually and would occur under hypothetical space. For example, if a customer bought a car with a discount, we will never know whether this customer would still buy that car if there were no discounts.

The vastly used solution to the unit selection problem is the A/B test heuristics, where a small set of individuals is divided into two groups: a control group and a treatment group. The treatment group is served with encouragement, whereas the control group is not. Individuals with certain characteristics that maximize the difference in effective rate are then selected as desired individuals. However, the counterfactual fact of the unit selection problem is not properly handled. In contrast, Li and Pearl labeled individuals into four counterfactual response types: complier, always-taker, never-taker, and defier, and proposed an objective function (i.e., benefit function) that is the average payoff of selecting each of the response types \cite{li:pea19-r488}. Using experimental and observational data, Li and Pearl provided the tight bounds of the benefit function and showed that their unit selection model is the right way to select the desired individuals compared to that of the A/B test heuristics.

In this study, we focus on the relationship between Li-Pearl's benefit function and the A/B test heuristics, explaining why the A/B test heuristics are sometimes problematic. Afterward, more simulated case studies are provided to emphasize the conclusion, as well as illustrate how to apply Li-Pearl's unit selection model correctly.

\section{Preliminaries}
\label{related work}
Here,  we review Li and Pearl's benefit function of the unit selection problem \cite{li:pea19-r488}. Readers who are familiar with the model may skip this section.

Similarly, we used the language of the structural causal model (SCM) \cite{galles1998axiomatic,halpern2000axiomatizing}, where the counterfactual sentences are well-defined. The experimental and observational data herein are those in the form of causal effects and a joint probability function denoted by $P(y_x)$ and $P(x,y)$, respectively. The basic counterfactual sentence ``Variable $Y$ would have the value $y$, had $X$ been $x$'' is denoted by $Y_x=y$. For simplicity, $Y_x=y$ is shorted as $y_x$. Besides, if not specified, the treatment is denoted by $X$, and the effect is denoted by $Y$.

% In this study we use the language of counterfactuals in structural model semantics, as given in \cite{galles1998axiomatic,halpern2000axiomatizing}. we use $Y_x=y$ to denote the counterfactual sentence ``Variable $Y$ would have the value $y$, had $X$ been $x$". For simplicity purposes, in the rest of the paper, we use $y_x$ to denote the event $Y_x=y$, $y_{x'}$ to denote the event $Y_{x'}=y$, $y'_x$ to denote the event $Y_x=y'$, and $y'_{x'}$ to denote the event $Y_{x'}=y'$. we assume that experimental data will be summarized in the form of the causal effects such as $P(y_x)$ and observational data will be summarized in the form of the joint probability function such as $P(x,y)$. If not specified, the variable $X$ stands for treatment and the variable $Y$ stands for effect.
Suppose the treatment is binary (denoted by $x$ and $x'$), and the effect is binary (denoted by $y$ and $y'$). Li and Pearl classified individual behaviors into four response types: complier, always-taker, never-taker, and defier. If the payoff of selecting an individual of each response type is $\beta, \gamma, \theta, \delta$ respectively (i.e., benefit vector), the objective function defined by Li and Pearl, which is the average payoff per selecting an individual, is as follows:
\begin{eqnarray}
\label{liobj}
argmax_c \text{ }\beta P(y_{x},y'_{x'}|c)+\gamma P(y_{x},y_{x'}|c) +\theta P(y'_{x},y'_{x'}|c)+\delta P(y'_{x},y_{x'}|c).
\end{eqnarray}
Note that $c$ represents the population-specific characteristics, and the benefit function is a linear combination of the probabilities of causation \cite{li:pea-r516, li2022bounds, pearl1999probabilities, tian2000probabilities}. Using a combination of experimental and observational data, Li and Pearl established the tight bounds of the above benefit function as follows (which we referred to as Li-Pearl's Theorem hereafter). The only assumption is that $c$ is not a descendant of treatment $X$.
\begin{theorem}
\label{thm1}
Given a causal diagram $G$ and distribution compatible with $G$, let $C$ be a set of variables that does not contain any descendant of $X$ in $G$, then the benefit function $f(c)=\beta P(y_x,y'_{x'}|c)+\gamma P(y_x,y_{x'}|c)+ \theta P(y'_x,y'_{x'}|c) + \delta P(y_{x'},y'_{x}|c)$ is bounded as follows:
\begin{eqnarray*}
W+\sigma U\le f(c) \le W+\sigma L\text{~~~~~~~~if }\sigma < 0,\\
W+\sigma L\le f(c) \le W+\sigma U\text{~~~~~~~~if }\sigma > 0,
\end{eqnarray*}
where $\sigma, W,L,U$ are given by,
\begin{eqnarray*}
&&\sigma = \beta - \gamma - \theta + \delta,W=(\gamma -\delta)P(y_x|c)+\delta P(y_{x'}|c)+\theta P(y'_{x'}|c),\\
&&L=\max\left\{
\begin{array}{c}
0,\\
P(y_x|c)-P(y_{x'}|c),\\
P(y|c)-P(y_{x'}|c),\\
P(y_x|c)-P(y|c)\\
\end{array}
\right\},
U=\min\left\{
\begin{array}{c}
P(y_x|c),\\
P(y'_{x'}|c),\\
P(y,x|c)+P(y',x'|c),\\
P(y_x|c)-P(y_{x'}|c)+\\+P(y,x'|c)+P(y',x|c)
\end{array}
\right\}.
\end{eqnarray*}
\end{theorem}
The above bounds have been narrowed by Li and Pearl \cite{li2022unit} given additional covariate information and the causal structure \cite{dawid2017, pearl:etal21-r505}. The nonbinary cases were discussed in ref \cite{li:pea22-r517}.

In addition, Li and Pearl provided conditions such that the benefit function can have a point estimation.
\begin{definition} (Monotonicity)
A Variable $Y$ is said to be monotonic relative to variable $X$ in a causal model $M$ iff
\begin{eqnarray*}
y'_x\land y_{x'}=\text{false}
\end{eqnarray*}
\label{df1}
\end{definition}
\begin{definition} (Gain Equality)
The benefit of selecting a complier ($\beta$), an always-taker ($\gamma$), a never-taker($\theta$), and a defier ($\delta$) is said to satisfy gain equality iff
\begin{eqnarray*}
\beta+\delta=\gamma+\theta
\end{eqnarray*}
\label{df2}
\end{definition}
\begin{theorem}
Given that $Y$ is monotonic relative to $X$ or that ($\beta, \gamma, \theta, \delta$) satisfies gain equality, the benefit function $f(c)$ is given by
\begin{eqnarray*}
f(c)=(\beta -\theta) P(y_x|c)+(\gamma-\beta) P(y_{x'}|c)+\theta
\end{eqnarray*}
\label{tm7}
\end{theorem}

\section{Essence of A/B Test Heuristic}
A common solution that is explored in the literature is an A/B test-based approach where a controlled experiment is performed, and the result is used as a selection criterion. Specifically, individuals are randomly split into two groups called control and treatment. Individuals in the control group are served with no treatment, whereas those in the treatment group are served with the treatment. Then, the commonly used A/B test heuristics are $a P(y_x|c) - bP(y_{x'}|c)$ (i.e., the weighted difference between the effective rates under treatment and no treatment). We then have, 
\begin{eqnarray*}
a P(y_x|c) - bP(y_{x'}|c) &=& a (P(y_x,{y'}_{x'}|c) + P(y_x,{y}_{x'}|c)) - b (P(y_{x'},{y'}_{x}|c) + P(y_{x'},{y}_{x}|c))\\
&=& a P(y_x,{y'}_{x'}|c) + (a-b)P(y_x,{y}_{x'}|c) - bP(y_{x'},{y'}_{x}|c).
\end{eqnarray*}
The effective rate under treatment is the percentage of the complier plus always-taker in the population, and the effective rate under no treatment is the percentage of always-taker plus defier in the population. The essence of the A/B test heuristics is a weighted difference between $P(complier\cup always\_taker)$ and $P(always\_taker\cup defier)$. Therefore, the A/B test heuristics are special cases of the benefit function. The benefit function has more expression power than the A/B test heuristics. This explains why the A/B test heuristics can be optimal for some cases (i.e., gain equality \cite{li:pea19-r488} satisfied.) and is generally problematic.

\section{Case Studies}
\label{chapter8}
Recall that the benefit vector in Li-Pearl's model is not determined by the model but by the model's user. Here, we illustrate several common applications showing how to set the benefit vector. The applications are categorized based on the quality of the A/B test-based approaches.

\subsection{Cases in which A/B-test Heuristics are Correct}
Here, we present simulated examples that satisfy the gain equality so that the benefit function is reduced to an A/B test Heuristics.

% \subsubsection{Number of Total Customers}
% If the manager only wants to maximize the total number of customers in the next quarter regardless of the total profit, then they should assign $1$ to a complier and an always-taker because the company has one customer in the next quarter and assign $0$ to a never-taker and a defier because the company has no customer in the next quarter.

% Therefore, the benefit vector above is $(1,1,0, 0)$, and using Theorem \ref{tm7}, when the benefit vector satisfies the gain equality ($1+0=1+0$), the benefit function is $f(c)=P(r_a|c)$. This is another common A/B test heuristic in literature, which is the causal effect of the offer to the number of customers. From the view of the essence of A/B test heuristics, $P(r_a|c)$ is exactly complier+always\_taker.

\subsubsection{Immediate Profit}

% If the manager wants to maximize the total immediate profit due to the offer. The management estimates that the benefit of selecting a complier is $\$100$ as the profit is $\$140$ but the discount is $\$40$, the benefit of selecting an always-taker is $-\$40$ as the customer would continue the service anyway and the company loses the value of the discount, the benefit of selecting a never-taker is $\$0$ as the cost of issuing the discount is negligible, and the benefit of selecting a defier is $-\$140$ as they lose a customer due to the special offer.

Consider a car manufacturer that wants to identify customers (based on observed characteristics) who would buy a hybrid car if there is an enticement and would not otherwise so as to increase their immediate total profit. 

The management of the car manufacturing plant acknowledged that the profit of sending the enticement to a complier (i.e., who would buy a hybrid car if there is an enticement and would not otherwise) is $\$45000$ as the profit of selling a hybrid car is $\$50000$, but the enticement costs $\$5000$. The profit of sending the enticement to an always-taker (i.e., who would buy a hybrid car no matter whether or not there is an enticement) is $-\$5000$ as the always-taker would buy the car anyway so that the manufacturer loses the enticement cost. The profit of sending the enticement to a never-taker (i.e., who would not buy a hybrid car no matter whether or not there is an enticement) is $\$0$ as the process of issuing the enticement is negligible. The profit of sending the enticement to a defier (i.e., who would buy a hybrid car if there is no enticement and would not otherwise) is $-\$50000$ as the manufacturer lost a customer due to the enticement.

% Therefore, the benefit vector above is $(100, -40,0, -140)$, using Theorem \ref{tm7}, when the benefit vector satisfies the gain equality ($100-140=-40+0$), the benefit function is $f(c)=100P(r_a|c)-140P(r_{a'}|c)$. This result is the same as the popular method in the industry, which is called revenue difference. The profit of a continuing customer if issued the special offer is $\$100$ and the profit of a continuing customer if no special offer is issued is $\$140$; therefore, the revenue difference is $100P(r_a|c)-140P(r_{a'}|c)$. From the view of the essence of A/B test heuristic, $100P(r_a|c)-140P(r_{a'}|c)$ is 100(complier+always\_taker)-140(always\_taker+defier)=100complier-40always\_taker-140defier.
The benefit vector above is $(45000, -5000,0, -50000)$, which satisfies the gain equality ($45000-50000=-5000+0$). The benefit function is then $f(c)=45000P(y_x|c)-50000P(y_{x'}|c)$. This is the same as the popular A/B test heuristics revenue difference in the industry. From the view of the essence of A/B test heuristic, $45000P(y_x|c)-50000P(y_{x'}|c)$ is $45000P(complier\cup always\_taker)-50000P(always\_taker\cup defier)=45000P(complier)-5000P(always\_taker)-50000P( defier)$.

\subsubsection{Number of Increased Customers}
Suppose the goal of the car manufacturer is to maximize the total increased customers regardless of the total profit. Therefore, for the benefit vector, $1$ should be assigned to the complier because the manufacturer gains one customer due to the enticement. Furthermore, $0$ should be assigned to both the always-taker and never-taker because the manufacturer gains no customer due to the enticement. Also, $-1$ should be assigned to the defier because the manufacturer loses one customer due to the enticement.
The benefit vector is then $(1,0,0, -1)$, which again satisfies the gain equality ($1-1=0+0$). The benefit function is $f(c)=P(y_x|c)-P(y_{x'}|c)$, which is another common A/B test heuristics. From the view of the essence of A/B test heuristic, $P(y_x|c)-P(y_{x'}|c)$ is $P(complier\cup always\_taker)-P(always\_taker\cup defier)=P(complier)-P(defier)$.

\subsection{Cases in which A/B-test Heuristics are Incorrect}

\subsubsection{Nonimmediate Profit}
Suppose the car manufacturer wishes to maximize the total profit but includes the long-term profit (e.g., sending enticement to an always-taker may change the customer to a complier). Therefore, the management should change the benefit of selecting an always-taker to $-\$7000$ (i.e., an estimated loss of $2000$ in the long term) in the example of immediate profit and keep the rest the same.

The benefit vector is $(45000, -7000,0, -50000)$. Then, the benefit function is $f(c)=45000P(y_x,{y'}_{x'}|c)-7000P(y_x,y_{x'}|c)-50000P({y'}_{x},y_{x'}|c)$, which no point estimation exists with pure experimental data. From the view of the essence of A/B test heuristics, no matter how to assign $a,b$ in $aP(y_x|c)-bP(y_{x'}|c)=aP(complier\cup always\_taker)-bP(always\_taker\cup defier)$, the results cannot equal to $45000P(complier) -7000P(always\_taker) -50000P(defier)$.

\subsubsection{Effectiveness of Vaccine}

A vaccine manufacturer invented a vaccine for a new virus and wants to verify the effectiveness of the vaccine. The effectiveness of a vaccine should be the difference between benefited and non-benefited (i.e., ineffective and harmed) individuals.

Let $X=x$ denote the event that an individual receives the vaccine, $X=x'$ denote the event that an individual receives no vaccine, $Y=y$ denote the event that an individual is not infected by the virus, $Y=y'$ denote the event that an individual is infected by the virus, and $C$ (a set of variables) denote the population-specific characteristics of an individual.

The commonly used way to assess the effectiveness of the vaccine is the A/B test heuristic, $P(y_x|c)-P(y_x'|c)$. However, this does not capture the counterfactual behavior. The real benefited individual should be the complier (i.e., the individual who would not be infected by the virus if vaccinated and would be infected if unvaccinated), the harmed individual should be the defier (i.e., the individual who would be infected by the virus if vaccinated and would not be infected if not vaccinated), and all other types (i.e., always-taker and never-taker) are all ineffective individuals. The benefit vector should then be $(1,-1,-1,-1)$, and the benefit function is $f(c)=P(y_{x},y'_{x'}|c) - P(y_{x},y_{x'}|c)-P(y'_{x},y'_{x'}|c)-P(y'_{x},y_{x'}|c)$. From the view of the essence of A/B test heuristics, $ P(y_x|c)-P(y_x'|c)$ is equal to $P(complier\cup always\_taker)-P(always\_taker\cup defier)=P(complier)-P(defier)$, which definitely cannot represent the desired objective function (i.e., $P(complier)-P(always\_taker)-P(never\_taker)-P(defier)$).

Let us plug some concrete numbers into the above example. Suppose two groups of individuals, $c_1$ and $c_2$, are waiting for the conclusion of whether the vaccine should be offered to the group. From the informer's view, two groups of individuals consist of different percentages of response types, as shown in Table \ref{tb1} (note that in the real world, we never know these numbers). The random sampled experimental data of $1500$ individuals of each group are given in Table \ref{tb2} (one can check that the simulated experimental data are compatible with the informer data, i.e., $P(y_x|c)=P(complier)+P(always\_taker)$ and $P(y_{x'}|c)= P(always\_taker)+P(defier)$). We also have the prior observational data that $P(y|x,c_1)=0.4$ and $P(y|x,c_2)=0.05$.

\begin{table}
\centering
\caption{Percentages of four response types in each group.}
\begin{tabular}{|c|c|c|c|c|}
\hline
&Complier&\begin{tabular}{c}Always-taker\end{tabular}&\begin{tabular}{c}Never-taker\end{tabular}&Defier\\
\hline
Group $c_1$&$35\%$&$25\%$&$35\%$&$5\%$\\
\hline
Group $c_2$&$65\%$&$5\%$&$5\%$&$25\%$\\
\hline
\end{tabular}
\label{tb1}
\end{table}

\begin{table}
\centering
\caption{Results of experimental samples on two groups.}
\begin{tabular}{|c|c|c|c|}
\hline
&&$do(x)$ & $do(x')$\\
\hline
{Group $c_1$}&$y$&450&225\\
&$y'$&300&525\\
\hline
{Group $c_2$}&$y$&525&225\\
&$y'$&225&525\\
\hline
\end{tabular}
\label{tb2}
\end{table}

The above-mentioned A/B test heuristic and the benefit function were compared, where,
\begin{eqnarray*}
f_1(c)=P(y_x|c)- P(y_{x'}|c).
\end{eqnarray*}
\begin{eqnarray*}
f_2(c)=P(y_{x},y'_{x'}|c) - P(y_{x},y_{x'}|c)-P(y'_{x},y'_{x'}|c)-P(y'_{x},y_{x'}|c).
\end{eqnarray*}

We plug the experimental and observational data into two objective functions, the results are summarized in Table \ref{tb3} (Midpoints of Theorem \ref{thm1} applied for $f_2$). The real values come from the informer data, where $real(c_1)=P(y_{x},y'_{x'}|c_1) - P(y_{x},y_{x'}|c_1)-P(y'_{x},y'_{x'}|c_1)-P(y'_{x},y_{x'}|c_1)=0.35-0.25-0.35-0.05=-0.3$ and $real(c_2)=P(y_{x},y'_{x'}|c_2) - P(y_{x},y_{x'}|c_2)-P(y'_{x},y'_{x'}|c_2)-P(y'_{x},y_{x'}|c_2)=0.65-0.05-0.05-0.25=0.3$. The conclusion from Theorem \ref{thm1} and $f_2$ is to offer the vaccine to group 2 only (i.e., the effectiveness is positive), which is the same as the real effectiveness from the informer's view. The conclusion from the A/B test heuristic is problematic in offering the vaccine to both groups though the real effectiveness of the first group is negative.

\begin{table}
\centering
\caption{Results of the two objective functions based on the data from the simulated study.}
\begin{tabular}{|c|c|c|c|}
\hline
&$f_1$&$f_2$&real\\
\hline
Group $c_1$&$0.3$&$-0.1$&$-0.3$\\
\hline
Group $c_2$&$0.4$&$0.35$&$0.3$\\
\hline
\end{tabular}
\label{tb3}
\end{table}

\subsubsection{Effectiveness of Vaccine with Focus on Effected Individual}
In the last example, the benefited individual, the harmed individual, and the ineffective individual is treated with the same importance. Suppose that we want to distinguish these individuals because we want to focus on benefited and harmed individuals; therefore, we could simply change the benefit vector to $(2,-1,-1,-2)$ to indicate such a requirement. 

The benefit function is then $f(c)=2P(y_{x},y'_{x'}|c) - P(y_{x},y_{x'}|c)-P(y'_{x},y'_{x'}|c)-2P(y'_{x},y_{x'}|c)$, which again has no point estimation with pure experimental data. From the view of the essence of A/B test heuristics, again, no matter how to setup $a,b$ in $aP(y_x|c)-bP(y_{x'}|c)=aP(complier\cup always\_taker)-bP(always\_taker\cup defier)$, it can never equal to the desired objective function $2P(complier)-P(always\_taker)-P(never\_taker)-2P(defier)$.

\section{Conclusion}
Herein, we reviewed Li-Pearl's unit selection model and its benefit function. We explained the relationships between the benefit function and the A/B test heuristics by showing that the essence of the latter one is a weighted difference between $P(complier\cup always\_taker)$ and $P(always\_taker\cup defier)$. Li and Pearl provided useful resources to bridge the theoretical unit selection model and the real-world applications, including the needed sample size and machine learning tools \cite{li:pea22-r519, li:pea22-r520, li:pea22-r518, li2020training}. We hope that the example provided here will help decision-makers to apply Li-Pearl's model correctly and push forward the revolution of A/B test heuristics style decision-making.

\section*{Acknowledgements}
This research was supported in parts by grants from the National Science
Foundation [\#IIS-2106908], Office of Naval Research [\#N00014-17-S-12091
and \#N00014-21-1-2351], and Toyota Research Institute of North America
[\#PO-000897].
\newpage
\bibliographystyle{plain}
\bibliography{neurips_2022.bib}

\appendix

% \section{Appendix}

% Optionally include extra information (complete proofs, additional experiments and plots) in the appendix.
% This section will often be part of the supplemental material.

\end{document}